\documentclass{article}

\usepackage[preprint]{neurips_2026}

\usepackage[utf8]{inputenc} 
\usepackage[T1]{fontenc}    
\usepackage{hyperref}       
\usepackage{url}            
\usepackage{booktabs}       
\usepackage{amsfonts}       
\usepackage{nicefrac}       
\usepackage{microtype}      
\usepackage{xcolor}         
\usepackage{amssymb}
\usepackage{amsmath}
\usepackage{xurl}
\usepackage[inline]{enumitem}
\usepackage{comment}
\usepackage{graphicx}
\usepackage{bbm}
\usepackage{algorithm}
\usepackage{algorithmic}
\usepackage{listings}
\usepackage[table]{xcolor}
\usepackage{caption}

\definecolor{light-gray}{gray}{0.95}

\definecolor{mygrey}{rgb}{0.94,0.96,0.96}

\colorlet{punct}{red!60!black}
\definecolor{background}{HTML}{EEEEEE}
\definecolor{delim}{RGB}{20,105,176}
\colorlet{numb}{magenta!60!black}

\definecolor{light-gray}{gray}{0.95}

\colorlet{punct}{red!60!black}
\definecolor{background}{HTML}{EEEEEE}
\definecolor{delim}{RGB}{20,105,176}
\colorlet{numb}{magenta!60!black}

\lstdefinelanguage{json}{
    basicstyle=\small\ttfamily,
    numbers=left,
    numberstyle=\scriptsize,
    stepnumber=1,
    numbersep=8pt,
    showstringspaces=false,
    breaklines=true,
    frame=lines,
    backgroundcolor=\color{background},
    literate=
     *{0}{{{\color{numb}0}}}{1}
      {1}{{{\color{numb}1}}}{1}
      {2}{{{\color{numb}2}}}{1}
      {3}{{{\color{numb}3}}}{1}
      {4}{{{\color{numb}4}}}{1}
      {5}{{{\color{numb}5}}}{1}
      {6}{{{\color{numb}6}}}{1}
      {7}{{{\color{numb}7}}}{1}
      {8}{{{\color{numb}8}}}{1}
      {9}{{{\color{numb}9}}}{1}
      {:}{{{\color{punct}{:}}}}{1}
      {,}{{{\color{punct}{,}}}}{1}
      {\{}{{{\color{delim}{\{}}}}{1}
      {\}}{{{\color{delim}{\}}}}}{1}
      {[}{{{\color{delim}{[}}}}{1}
      {]}{{{\color{delim}{]}}}}{1},
}

\title{End-to-end PDDL Planning with Hardcoded and Dynamic Agents}

\author{%
  Emanuele La Malfa$^{1,3}$\thanks{Equal contribution. Keep the correspondence to \texttt{emanuele.lamalfa{@}cs.ox.ac.uk}.} \quad 
  Ping Zhu$^{1}$\textsuperscript{$*$} \quad 
  Samuele Marro$^{2,3}$ \\
  \textbf{Sara Bernardini}$^{1}$ \quad
  \textbf{Michael Wooldridge}$^{1}$ \\
  $^{1}$Department of Computer Science, University of Oxford \\
  $^{2}$Department of Engineering, University of Oxford \\
  $^{3}$IDAI
}

\begin{document}

\maketitle

\vspace{-0.8cm}
\begin{center}
    \includegraphics[height=1em]{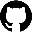} \href{https://github.com/EmanueleLM/MultiAgentPlanning}{[\texttt{Code \& Results}]}
\end{center}

\begin{abstract}
    We present an end-to-end framework for planning supported by verifiers. An orchestrator receives a human specification written in natural language and converts it into a PDDL (Planning Domain Definition Language) model, where the domain and problem are iteratively refined by sub-modules (agents) to address common planning requirements, such as time constraints and optimality, as well as ambiguities and contradictions that may exist in the human specification. We support two categories of agents: hardcoded, which are informed by logs and error traces and have a pre-defined goal (e.g., fix issues with PDDL syntax, check temporal constraints), and dynamic, which have no predefined goal but adapt to the specific domain and revise the latent planning abstraction. The validated domain and problem are then passed to an external planning engine to generate a plan. The orchestrator and agents are powered by Large Language Models (LLMs) and require no human intervention at any stage of the process. Finally, a module translates the final plan back into natural language to improve human readability while maintaining the correctness of each step. We demonstrate the flexibility and effectiveness of our framework on GPT-\{4o, 5-mini, 5.4\}, and Gemini-\{2.5, 3\}-flash across more than ten domains and tasks, including the Google NaturalPlan benchmark, Planbench, and classic planning problems like Sokoban, Blocksworld and the Tower of Hanoi, where LLMs are known to struggle even with small instances. Our framework can be integrated with any PDDL planning engine and validator (we successfully tested Fast Downward, LPG, POPF, VAL, and uVAL) and represents a significant step toward end-to-end planning aided by LLMs.
\end{abstract}

\section{Introduction}
Planning is a cornerstone of intelligent behaviour: from robotics and logistics to digital assistants and collaborative environments, agents must translate high-level goals into structured sequences of actions. Automated planning, one of the most longstanding subfields of AI, offers guarantees of correctness and optimality \cite{fikes1971strips,ghallab1998pddl}. However, it requires expert-crafted domain models, typically in PDDL (Planning Domain Definition Language) \cite{McDermott1998PDDL}, and is ill-suited to handle ambiguous, incomplete, or underspecified requirements expressed in natural language. In contrast, Large Language Models (LLMs) excel at interpreting complex instructions but struggle to produce long-horizon or multi-agent plans directly, often hallucinating steps or producing infeasible sequences \cite{valmeekam2023planbench,zuo2024planetarium}.

Recent work has sought to bridge these paradigms by using LLMs as front-ends to symbolic solvers, translating natural language descriptions into structured representations \cite{ahn2022can,gundawar2024robust}. While such pipelines reduce the modelling burden, they typically rely on fixed, pre-defined workflows. Multi-agent systems allow for planning with shared resources, asynchronous execution, and dynamic coordination \cite{lowe2017multi,oliehoek2016concise}. Addressing these challenges requires systems that are simultaneously linguistically capable, formally grounded, and robust to diverse scenarios.

\begin{figure*}
\centering
\includegraphics[width=1\textwidth]{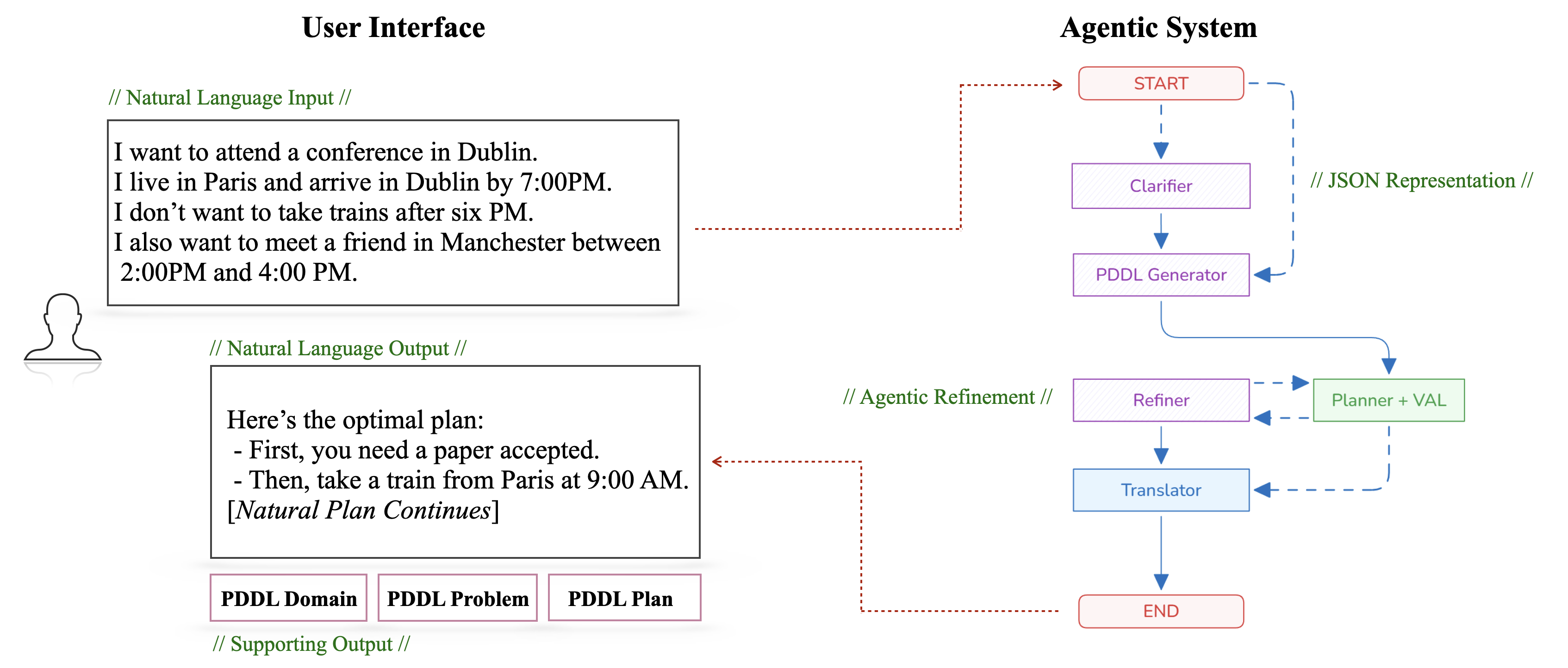} 
\caption{
Overview of our end-to-end planning framework graph that generates a plan, backed up by a planner, from a human specification (left). The framework includes the first JSON and PDDL plan generator, and the agentic sub-component, which refines an existing plan (right).}\label{fig:intro}
\end{figure*}

In this paper, we present the first end-to-end agentic framework for planning that unifies natural language understanding, dynamic agent orchestration, symbolic reasoning, and plan interpretation (Figure~\ref{fig:intro}). Unlike prior approaches with static role assignment \cite{erdogan2025plan}, our orchestrator LLM dynamically creates multi-agent workflows on the fly, asynchronously invoking agents to construct PDDL domains and problems, which are first merged, then iteratively refined by a sequence of specialised agents depending on the task at hand. The refined PDDL problem is then solved by a planner to guarantee correctness with respect to the specification and, potentially, cost-optimality. The resulting plan is automatically translated back into natural language, which ensures accessibility and interpretability while preserving formal correctness. Our contribution is threefold:
\begin{itemize}
    \item We introduce a fully automated, end-to-end multi-agent planning pipeline that closes the gap between natural language intent communication and optimal plans, enabling the system to move from free-form specifications to validated plans with no human intervention. We show how hardcoded agents can mitigate some longstanding obstacles in planning, including ambiguity resolution in a human specification~\cite{grosz1996collaborative, tambe1997towards}. On the other hand, dynamic agents enhance the flexibility of planning workflow generation.
    \item We run experiments on more than ten benchmarks that cover long-horizon planning tasks and constraint satisfaction problems (CSP). On the former, our approach outperforms vanilla LLMs and has nearly perfect success rate. On the latter, the gap with vanilla models remains large for non-frontier models such as GPT-5-mini and Gemini-2.5 but reduces on models like GPT-5.4. Also, our method solves hard instances of the Tower of Hanoi and Blocksworld, where LLMs are known to struggle~\cite{monti2026sokobenchevaluatinglonghorizonplanning,shojaee2025illusion}.
    \item Our work identifies CSP as the setting where PDDL+LLMs (but, in general, neuro-symbolic techniques) still struggle. It is also a call for action for the research community to carefully develop evaluation techniques to mitigate the brittleness of LLMs as-a-judge when the ground truth label and the final plan are expressed in natural language.  
\end{itemize}

Overall, our framework represents a step toward unifying LLMs and symbolic planning into general-purpose autonomous systems. By dynamically orchestrating multiple agents, grounding reasoning in solvers, and ensuring human-auditable outputs through natural language back-translation, we push the frontier of end-to-end multi-agent planning toward real-world applicability. This positions our approach as both a practical and conceptual advancement in the integration of language models and multi-agent symbolic reasoning.

\section{Background and Related Work}
We provide an overview of research at the intersection of LLMs, automated planning, and agentic systems, highlighting key trends in the literature.

\paragraph{LLMs for planning.}
Early attempts to use LLMs for planning focused on directly generating action sequences from natural-language task descriptions. However, studies show that LLMs alone are brittle, with even modern models exhibiting subpar performance \cite{gundawar2024robust}. In fact, a line of works \cite{Raollmscantplan, stillcantplan} reinforces that while LLMs by themselves are poor planners, they perform notably better when augmented with symbolic components. Valmeekam \cite{selfcritique} also underlines that without extra tooling, LLMs alone are unreliable at reasoning about their planning decisions (a pitfall that our system circumvents by introducing external feedback). More robust frameworks integrate LLMs with symbolic solvers. In \textit{LLM+P} pipeline~\cite{liu2023llmp}, LLMs attempt to directly translate language into PDDL specifications and subsequently call planners. Other approaches integrate LLMs with search algorithms: \textit{SayCan}~\cite{ahn2022can} uses LLMs with value functions to ground high-level instructions. Similarly,  \textit{SayCanPay}~\cite{hazra2024saycanpay} introduces coupling an LLM with A* search to guarantee cost-feasible plans. 
Another promising line treats planning as a form of program synthesis. Silver~\cite{silver2024generalized} shows the potential of GPT-4 as a generalised planner by synthesising Python programs that generate valid plans for PDDL domains, using iterative prompting and automated debugging loops to improve reliability. Mahdavi~\cite{mahdavi2024leveragingenvironmentinteractionautomated} leverages LLMs to generate valid PDDL domains and problems that are then fed to a solver, but requires access to the model's logprob, which is not available in any recent API-based LLM.
Overall, the literature shows a shift from using LLMs on their own to hybrid systems that combine the generative flexibility of LLMs with the correctness guarantees of symbolic planners.

\paragraph{Multi-agent and agentic systems.}
Modern research on multi-agent planning emphasizes learning-based coordination. Lowe~\cite{lowe2017multi} proposed a multi-agent actor-critic algorithm that addresses non-stationarity in concurrent learning via centralised critics. Differentiable communication channels \cite{foerster2016learning,sukhbaatar2016learning} further enable emergent coordination strategies. Benchmarks such as MeltingPot~\cite{leibo2021scalable} and MPE~\cite{lowe2017multi} evaluate cooperative and competitive multi-agent behaviours.
Additionally, recent work integrates LLMs as agents in collaborative environments. For example, Wu~\cite{wu2024your} introduced \textit{CoBlock}, a collaborative Blocksworld domain where two LLM-driven agents coordinate on construction tasks. The results show that incorporating chain-of-thought reasoning into agent prompts significantly improved coordination success rates and ensured more balanced workloads. However, prompts still included the world state at each decision iteration, reinforcing the challenges current LLMs face when  holding complex states in persistent memory and underscoring the benefit of outsourcing this task to a symbolic engine.

\paragraph{Automated planning and PDDL.}
Automated planning has long served as the formal backbone for generating autonomous behaviour. The STRIPS formalism \cite{fikes1971strips} laid the foundation for domain-independent planning, later standardised in PDDL \cite{McDermott1998PDDL}. Planners such as GraphPlan \cite{blum1997fast}, FF \cite{hoffmann2001ff}, and Fast Downward \cite{helmert2006fast} achieve efficient heuristic search across domains. However, designing correct PDDL models remains a bottleneck in the widespread adoption of verifiable planning. Recent work leverages LLMs to automate or assist with this process. \cite{zuo2024planetarium} show that LLMs are often better at generating formal specifications (in PDDL) than directly producing valid plans.  Additionally, benchmarks such as Planbench \cite{valmeekam2023planbench} highlight that, while LLMs struggle to outperform solvers, they are promising for lowering the modelling barrier. Oswald~\cite{oswald2024large} explores using GPT-4 to generate entire PDDL domains, though the proposed system still requires iterative refinement.

\paragraph{Tools, benchmarks, and datasets.}
A wave of benchmarks and open-source resources has supported modern evaluation of multi-agent planning in the context of LLMs. Zheng~\cite{zheng2024natural} introduced \textit{NATURAL PLAN}, a benchmark for realistic language-based planning tasks (e.g., trip planning, scheduling).
Planbench \cite{valmeekam2023planbench} offers standardised domains and metrics for LLM planning, while Planetarium \cite{zuo2024planetarium} provides a large dataset (145k pairs) for text-to-PDDL evaluation with equivalence checking tools.
For multi-agent collaboration, CoBlock \cite{wu2024your} explicitly tests human-LLM and LLM-LLM collaboration in Blocksworld scenarios. Embodied benchmarks such as ALFRED \cite{shridhar2020alfred} and its symbolic variant ALFWorld \cite{shridhar2020alfworld} bridge natural language instructions with embodied planning. Other environments, including BabyAI \cite{babyai}, remain popular for testing instruction following and sequential decision-making.

\section{Agentic End-to-End Planning}
In this section, we introduce planning as an optimisation problem, and the architecture to solve, in an end-to-end fashion, a wide range of planning tasks. We assume there is a specification and a ground truth plan, both written in natural language, which is the domain where our framework operates.
Our notation does not conflict but extends that of Mahdavi~\cite{mahdavi2024leveragingenvironmentinteractionautomated} as it assumes one can either verify the final PDDL plan (if any) or the natural language representation of it.

\subsection{Planning as an Optimisation Problem}
A planning task consists of a specification $s$, i.e., the description of the problem, and its ground truth solution $y$; we assume that both $s$ and $y$ live in $\Sigma^*$, which denotes the Kleene closure of all the possible strings in an alphabet of symbols, including natural language and PDDL specifications. 

Our agentic framework $F$ receives as input a task $s$ and produces a candidate solution for the task, which consists of a PDDL domain and problem $(d, p)$, namely $F: s \xrightarrow{} (d, p)$. 
We denote with $V$ a solver that turns the PDDL domain and problem into a candidate plan, i.e., $\hat{y} = V(d, p)$.
While $F$ iteratively selects the best candidate from a finite pool of agents $A = \{a_1, ..., a_n\}$ to produce the new candidate PDDL domain and problem, $V$ produces the PDDL plan and can be any PDDL solver.

While for some tasks the correctness of a candidate solution can be judged by comparing two PDDL plans (e.g., the Tower of Hanoi), in a generic scenario the ground truth plan is expressed in natural language: in both cases, we rely on an oracle that scores a candidate solution $\hat{y}$ (or $\hat{y}_{nl}$) as right ($1$) or wrong ($0$) against $y$, i.e., $\Omega: (\hat{y}, y) \xrightarrow{} \{0, 1\}$.\footnote{A formulation that allows scores to live in $[0, 1] \subset \mathbb{R}$ would be mathematically equivalent in our framework.}

For a pair of problem specifications and a ground truth solution sampled from a distribution $T$, namely $(s, y) \sim T$, our framework optimises the following:
\begin{equation}\label{eq:planning}
\begin{aligned}
\max_{F} \quad & \mathbb{E}_{(s,y) \sim T} [\Omega(\hat{y}_{nl}, y)] \\
\textrm{s.t.} \quad & \hat{y}_{nl} = (V \circ F(s))^{\circ b} \\
\quad & b \in \mathbb{N}_{> 0}, \\
\end{aligned}
\end{equation}

where $(V \circ F(s))^{\circ b}$ denotes how our framework $F$ alternates with the solver $V$ to produce the candidate PDDL plan, while consuming a \emph{budget} $b > 0$.
In other words, we look for the agentic system $F$ that optimises the choice of agents at each substep (more details in the next section).
The optimisation problem outlined above pertains to the case where the ground truth plan is expressed in PDDL. W.l.o.g., the extension to natural language involves a translation module $Nl$ (an LLM agent) that turns the candidate plan $\hat{y}$ into natural language, i.e., $\hat{y}_{nl} = Nl \circ (V \circ F(s))^{\circ b}$.

\subsection{Planning as a Sequential Agentic Problem}
Our design of Eq.~\ref{eq:planning} and function $F$ are inspired by recent advances in agentic LLMs~\cite{chan2024mle,toledo2025ai}, which combine the ability of LLMs to identify which parts of a plan should be refined with the formality of PDDL solvers and validators.

Our process begins by instantiating an orchestrator agent that generates a structured representation of the environment, the agents, their goals, and their constraints from the natural language task description. We adopt JSON as the intermediate representation because it flexibly captures heterogeneous data and is widely represented in open- and closed-source LLM pre-training corpora~\cite{shorten2024structuredrag}.
The orchestrator then produces an initial PDDL domain and problem, which are passed to a solver (e.g., Fast Downward or POPF2) to compute an initial candidate plan.

\begin{figure*}
\centering
\includegraphics[width=0.85\textwidth]{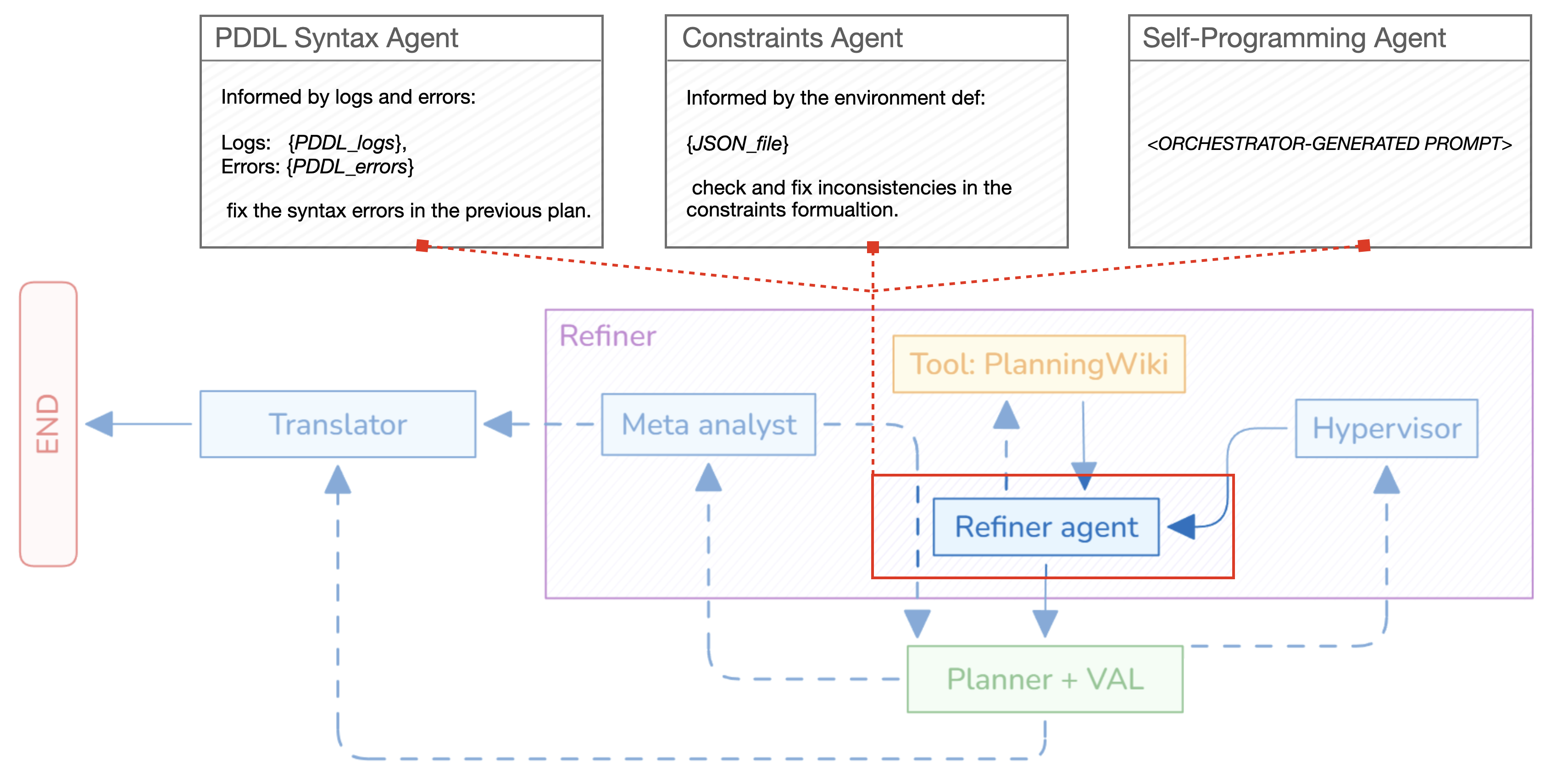} 
\caption{Overview of our planning framework illustrating how the hardcoded and dynamic agents interact with each other. The ``Refiner'' block can call agents whose prompt is pre-defined and solves specific issues with the current planning task (e.g., PDDL syntax errors), or generates a prompt for a dynamic agent that adapts to the current context.}\label{fig:overview}
\end{figure*}

The domain and problem are then iteratively refined. Using the logs produced by an external PDDL validation tool (VAL/uVAL), the orchestrator selects the most suitable agent from a pool of available agents. Some agents are hardcoded, i.e., they provide fixed procedures to repair PDDL syntax errors, adapt the domain and problem to the target solver, generate the final natural language plan, or terminate the computation early. In addition, we introduce an agent with no predefined prompt, which adapts to the current execution state. Figure~\ref{fig:overview} illustrates the agent selection process.

With a finite pool of hardcoded agents $A=\{a_1,\dots,a_n\}$, the pointwise objective in Eq.~\ref{eq:planning} can be rewritten as
\begin{equation}
a^\star \in \arg\max_{a \in A} \;
\mathbb{E}_{(s,y)\sim T}\!\left[
\Omega\!\bigl(\hat{y}_{nl}(s;a), y\bigr)
\right],
\end{equation}
where $\hat{y}_{nl}(s;a)$ denotes the candidate solution obtained when agent $a$ is selected. This corresponds to a standard multi-armed bandit problem, in which each agent is an arm with unknown expected reward.

When the orchestrator can choose between hardcoded and dynamic agents, the problem becomes
\begin{equation}
a^\star \in \arg\max_{a \in \mathcal{A}} \;
\max_{\theta \in \Theta_a}
\mathbb{E}_{(s,y)\sim T}\!\left[
\Omega\!\bigl(\hat{y}_{nl}(s;a,\theta), y\bigr)
\right],
\end{equation}
where $\Theta_a \subseteq \Sigma^*$ denotes the configuration space of agent $a$, such as prompts or generated programs.

This objective selects the agent $a \in \mathcal{A}$ with the highest achievable task performance, where performance is measured by the number of correctly solved instances (success rate). For each agent, an inner optimisation is carried out over its configuration space $\Theta_a$, yielding the best performance attainable by that agent under an optimal configuration. Hence, the outer optimisation selects the agent, while the inner optimisation searches over its configurations. This changes the nature of the problem: instead of selecting among a fixed set of arms, one must solve a potentially high-dimensional and combinatorial optimisation problem for each agent. Consequently, the main difficulty shifts from statistical estimation to structured search, and the resulting formulation no longer reduces to a standard bandit problem.

In the Appendix, we comprehensively discuss the orchestrator, the hardcoded and dynamic agents, and include their prompts. We also include the structure and an example of a JSON entirely generated by GPT-5-mini, and document how to use a web-demo of our tool.

\section{Experimental Evaluation}
\begin{table}
    \resizebox{\linewidth}{!}{
        \begin{tabular}{lcccc|ccc}
        \toprule
        \textbf{} 
        & \multicolumn{4}{c|}{{\textbf{Planbench}}} 
        & \multicolumn{3}{c}{{\textbf{Google Natural Plan}}} \\
        \textbf{} 
        & {\small\textbf{\shortstack{Depots \\ \hspace{0cm}}}} 
        & {\small\textbf{\shortstack{Logistics \\ \hspace{0cm}}}} 
        & {\small\textbf{\shortstack{Mystery \\ Blocksworld}}} 
        & {\small\textbf{\shortstack{Obfuscated Dec. \\ Logistics}}} 
        & {\small\shortstack{\textbf{Calendar} \\ \textbf{Scheduling}}} 
        & {\small\textbf{\shortstack{Meeting \\ Planning}}} 
        & {\small\textbf{\shortstack{Trip \\ Planning}}} \\
        \midrule
        GPT-5.4 & & & & & & & \\
        \quad + \textsc{Vanilla} & $21.3\pm1.8$ & $49.3\pm2.4$ & $33.3\pm2.4$ & $7.3\pm2.4$ & $82.9\pm2.2$ & $85.3\pm2.4$ & $5.3\pm0.9$ \\
        \quad + \cellcolor{mygrey}\textsc{Agents (Ours)} & \cellcolor{mygrey}$90.0\pm4.3$ & \cellcolor{mygrey}$100.0\pm0.0$ & \cellcolor{mygrey}$100.0\pm0.0$ & \cellcolor{mygrey}$100.0\pm0.0$ & \cellcolor{mygrey}$85.3\pm0.9$ & \cellcolor{mygrey}$81.3\pm3.4$ & \cellcolor{mygrey}$3.3\pm2.4$ \\
        \hline
        Gemini-3-flash & & & & & & & \\
        \quad + \textsc{Vanilla} & $99.2\pm0.9$ & $96.8\pm1.4$ & $97.0\pm2.0$ & $97.5\pm1.4$ & $93.0\pm1.5$ & $84.4\pm4.1$ & $98.4\pm1.6$ \\
        \quad + \cellcolor{mygrey}\textsc{Agents (Ours)} & \cellcolor{mygrey}$99.0\pm1.0$ & \cellcolor{mygrey}$97.5\pm1.7$ & \cellcolor{mygrey}$97.3\pm1.8$ & \cellcolor{mygrey}$97.3\pm0.9$ & \cellcolor{mygrey}$93.0\pm1.7$ & \cellcolor{mygrey}$84.0\pm4.5$ & \cellcolor{mygrey}$98.0\pm2.0$ \\
        \bottomrule
        \end{tabular}
    }
    \caption{Success rate (\%) of vanilla LLMs vs our agentic technique for the Planbench and Google Natural Plan benchmark. Results are averaged on three independent runs with $50$ problems each.}
    \label{tab:google-planbench}
\end{table}

We evaluate our approach on ten planning tasks: the Google Natural Plan benchmark~\cite{zheng2024natural} (calendar scheduling, meeting planning, trip planning), Planbench~\cite{valmeekam2023planbench} (depots, logistics, mystery blocksworld, obfuscated deceptive logistics), Tower of Hanoi and Blocksworld at varying complexity, and Sokoban~\cite{monti2026sokobenchevaluatinglonghorizonplanning}.
All experiments use API-accessible LLMs: GPT-5-mini, GPT-5.4, Gemini-2.5-flash, and Gemini-3-flash. We set the budget $b$ for plan generation and refinement to four and detail, in Section~\ref{sec:discussion} how many calls we require on average. We also compare our method with Exploration Walk~\cite{mahdavi2024leveragingenvironmentinteractionautomated} on two datasets where it achieves zero success (floortile and childsnack). 

We measure, for each task, the success rate of our approach and compare it to the same \emph{vanilla} LLM that receives the natural language prompt and returns the plan. We employ GPT-5-mini as-a-judge that evaluates whether the natural language plan generated by a method corresponds to the ground truth label (also expressed in natural language).
We first report the results on Planbench, which is a classic long-horizon planning benchmark adapted to test LLMs planning capabilities.
We then focus on the Google Natural Plan benchmark, which tests the capabilities of a model to understand the input question and perform constraint satisfaction.
We then discuss the results on Hanoi, Sokoban, and Blocksworld, which are similar to Planbench but are often used to show that LLMs cannot plan~\cite{shojaee2025illusion}, even when equipped with PDDL~\cite{monti2026sokobenchevaluatinglonghorizonplanning}.

We also do an ablation of our system where we remove the dynamic agent, to understand the contribution of each hardcoded agent.
We then discuss some open problems in planning with LLMs: in particular, additional experiments open research opportunities to mitigate the brittleness of LLMs as-a-judge, i.e., evaluating the correctness of a natural plan when no ground truth plan is available.

\subsection{Planbench}
Planbench~\cite{valmeekam2023planbench} is an extensible benchmark suite for long-horizon planning, i.e., the solution consists of a series of actions executed sequentially to bring the environment from an initial state, to the desired final configuration. In our evaluation, we test the depots, logistics, mystery blocksworld, and obfuscated deceptive logistics datasets.

As Table~\ref{tab:google-planbench} shows (left), on GPT-5.4 our approach yields a perfect success rate ($100\%$) on three out of four tasks, and more than $90\%$ on depots. By contrast, the vanilla model performs much worse, showing that our approach substantially improves the planning capabilities of GPT-5.4. With Gemini-3-flash, the vanilla and the PDDL approach achieve comparable, near-perfect success rates, with an average of $97\%$ and a negligible gap of $0.1\%$ between the two methods (the standard deviation across task success rates is $1\%$ for both methods).

Our approach is particularly effective with smaller models, achieving an average success rate on Planbench of $91.6\pm1.6$ for GPT-5-mini, compared to $76.6\pm23.2$ for the vanilla model. Similarly, on Gemini-2.5-flash, the average success rate on Planbench for our method is $89.7\pm6.8$, compared to $74.7\pm14.9$ for the vanilla model.

\subsection{Google Natural Plan Benchmark}
The Google Natural Plan benchmark~\cite{zheng2024natural} is a widely used task for evaluating the planning capabilities of LLMs, and consists of three tasks: calendar scheduling, where individuals must find a suitable slot in their calendars to meet; meeting planning, where a person travels from different locations and aims to meet a friend who is available only at a specific time; and trip planning, where a person wishes to visit multiple locations, spend a certain number of days in each, and can travel between locations only if a direct flight exists.
While the benchmark suggests that these tasks involve planning, they are more akin to constraint satisfaction problems (CSP)~\cite{amonkar2026realitychecklanguagemodels,katz2025makeplanningresearchrigorous}; therefore, they assess different capabilities from those examined in the following sections.

As Table~\ref{tab:google-planbench} reports (right), our method matches the performance of the vanilla GPT-5.4 and Gemini-3-flash: differently from Planbench, we observe that this testbed is already saturated by the base models; that alongside the difference in nature between Planbench and the Google Natural Plan (one is long-horizon planning, the other is CSP), probably explain why our approach does not outperform large frontier LLMs.

On the other hand, our approach dramatically increases the performance of GPT-5-mini, with our approach that has a success rate of $93.3\%$, $53.3\%$, and $8.0\%$ (average success rate: $51.5\pm34.8$) on the three reference tasks (calendar scheduling, meeting planning, and trip planning), compared to $88\%$, $24\%$, and $2\%$ (average success rate: $38.0\pm36.4$) of the vanilla model. 

\subsection{Comparison to Existing LLM + PDDL Planning Techniques}
\noindent
\begin{minipage}[t]{0.53\textwidth}
\vspace{0pt}
We compare our approach with Exploration Walk~\cite{mahdavi2024leveragingenvironmentinteractionautomated}, an LLM+PDDL planning technique that uses logprob inspection to provide rich feedback signals for LLMs to update the PDDL file.
We select two tasks where their method achieves a $0\%$ success rate: floortile,  which involves multiple robots navigating a grid of floor tiles to paint them in specific patterns using different colors, and childsnack, which asks how to make and serve sandwiches for a group of children, some of whom are allergic to gluten.
Floortile is a long-horizon planning problem, while childsnack is closer to CSP. Using the same model as in their experiments, i.e., GPT-4o, we achieve a $43.7\%$ success rate on floortile, while matching their $0\%$ success rate on childsnack. In this sense, the results reflect the duality between Planbench and Google Natural Plan: PDDL helps LLMs on planning, while it reduces effectiveness on CSP tasks.
While Exploration Walk cannot be applied to models such as GPT-5-mini or Gemini-2.5-flash (or newer), since it requires access
\end{minipage} \hspace{0.2cm}
\begin{minipage}[t]{0.43\textwidth}
\vspace{0pt}
\centering
\small
\setlength{\tabcolsep}{4pt}
\begin{tabular}{lcc}
\toprule
& \textbf{Floortile} & \textbf{Childsnack} \\
\midrule
\multicolumn{3}{l}{\textbf{GPT-4o}} \\
\quad \textsc{Vanilla} & $0.$ & $0.$ \\
\quad \textsc{Expl. Walk} & $0.$ & $0.$ \\
\quad \cellcolor{mygrey}\textsc{Agents (Ours)} & \cellcolor{mygrey}$43.7$ & \cellcolor{mygrey}$0.$ \\
\midrule
\multicolumn{3}{l}{\textbf{GPT-5-mini}} \\
\quad \textsc{Vanilla} & $26.0$ & $0.$ \\
\quad \textsc{Expl. Walk} & N/A & N/A \\
\quad \cellcolor{mygrey}\textsc{Agents (Ours)} & \cellcolor{mygrey}$68.7$ & \cellcolor{mygrey}$34.0$ \\
\midrule
\multicolumn{3}{l}{\textbf{Gemini-2.5-flash}} \\
\quad \textsc{Vanilla} & $10.5$ & $5.2$ \\
\quad \textsc{Expl. Walk} & N/A & N/A \\
\quad \cellcolor{mygrey}\textsc{Agents (Ours)} & \cellcolor{mygrey}$53.8$ & \cellcolor{mygrey}$53.3$ \\
\bottomrule
\end{tabular}
\captionof{table}{Success rates (\%) of vanilla LLMs, Exploration Walk~\cite{mahdavi2024leveragingenvironmentinteractionautomated}, and our agents on floortile and childsnack. E. Walk works with GPT-4o, but not for GPT-5, Gemini-2.5 or newer, as they do not return the logits.}\label{tab:ewalk-comparison}
\end{minipage}

to the logits to build the trace of executability of random action sequences sampled from the domain,
we compare the performance of frontier models with and without our approach. Vanilla GPT-5-mini achieves success rates of $26\%$ and $0\%$, whereas our approach achieves $68.7\%$ and $34\%$. 
Similarly, Gemini-2.5-flash achieves $10.5\%$ and $5.2\%$ without our method, versus $53.8\%$ and $53.3\%$ with it. 
Table~\ref{tab:ewalk-comparison} reports the results for GPT-4o, GPT-5-mini, and Gemini-2.5-flash.

\begin{figure*}[b]
\centering
\includegraphics[width=1\textwidth]{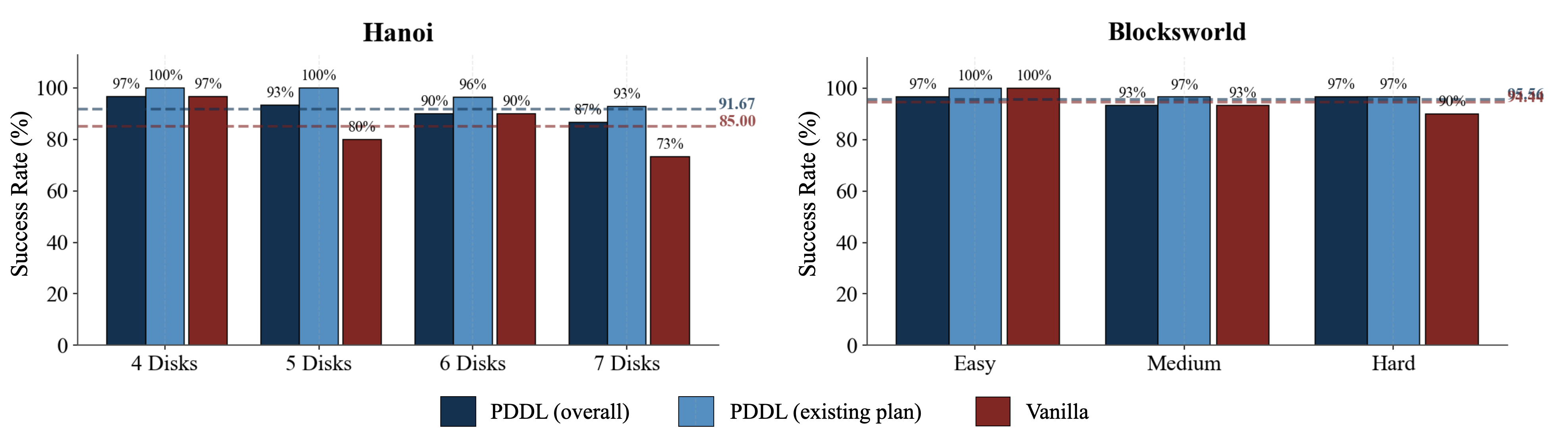} 
\caption{Success rate of our technique vs vanilla GPT-5-mini for increasingly complex Hanoi and blocksworld tasks. For our method, we report the overall success rate and the success rate when a plan is generated.}\label{fig:gpt-mini-hanoi-blocksworld}
\end{figure*}

\subsection{Hanoi, Blocksworld, and Sokoban: the Delusion of the Illusion of Thinking}\label{sec:discussion-hanoi}
We further assess the planning capabilities of our method on Blocksworld and the Tower of Hanoi, two standard benchmarks in planning. We sample several Blocksworld and Tower of Hanoi problems with increasing complexity (i.e., the optimal number of steps to solve an instance, which we enforce as part of the instructions to the model), and compare the success rate of our method with that of the same vanilla LLM (GPT-5-mini).
As shown in Figure~\ref{fig:gpt-mini-hanoi-blocksworld}, on the Tower of Hanoi, where recent influential papers show that LLMs cannot solve instances larger than 6 disks~\cite{valmeekam2023planbench}, we achieve $91\%$ of success rate ($+6\%$ on average), compared to $73\%$ ($+14\%$) of vanilla GPT-5-mini.
On Blocksworld with solutions that require exactly $2 - 4$, $6 - 8$, and $10 - 12$ actions per problem, which we denote as “Easy”, “Medium”, and “Hard”, our framework pairs the performance of GPT-5-mini on “Easy” and “Medium” instances; yet, it consistently outperforms it on “Hard” problems ($+7\%$). Our results reinforce the belief that LLMs can solve the hard instances of the Hanoi tower~\cite{lawsen2025commentillusionthinkingunderstanding}, contradicting influential papers in the area~\cite{shojaee2025illusion} (which only report results for Claude models).

On Sokoban, we focus only on the hard scenarios described in~\cite{monti2026sokobenchevaluatinglonghorizonplanning} (i.e., the length of the \emph{corridor} is between $90$ and $100$), as their results already show that LLMs perform very well in the other cases. While they report a success rate for GPT-5-mini of $\sim 13\%$ (we averaged their performance over our interval of corridor lengths), our method achieves $34\%$. GPT-5 has a vanilla success rate of $34\%$, and $40\%$ with our method. On the other hand, vanilla Gemini-\{2.5, 3\}-flash have a success rate of $78\%$ and $76\%$, while with our technique it drops to achieve $20\%$ and $68\%$, respectively.

In conclusion, the experiments in these sections depict a composite scenario where LLMs, with and without PDDL, succeed on ``standard'' planning tasks but fail on CSP. We discuss this and other open problems in planning with LLMs in Section~\ref{sec:discussion} and the conclusions.

\begin{figure*}
\centering
\includegraphics[width=1\textwidth]{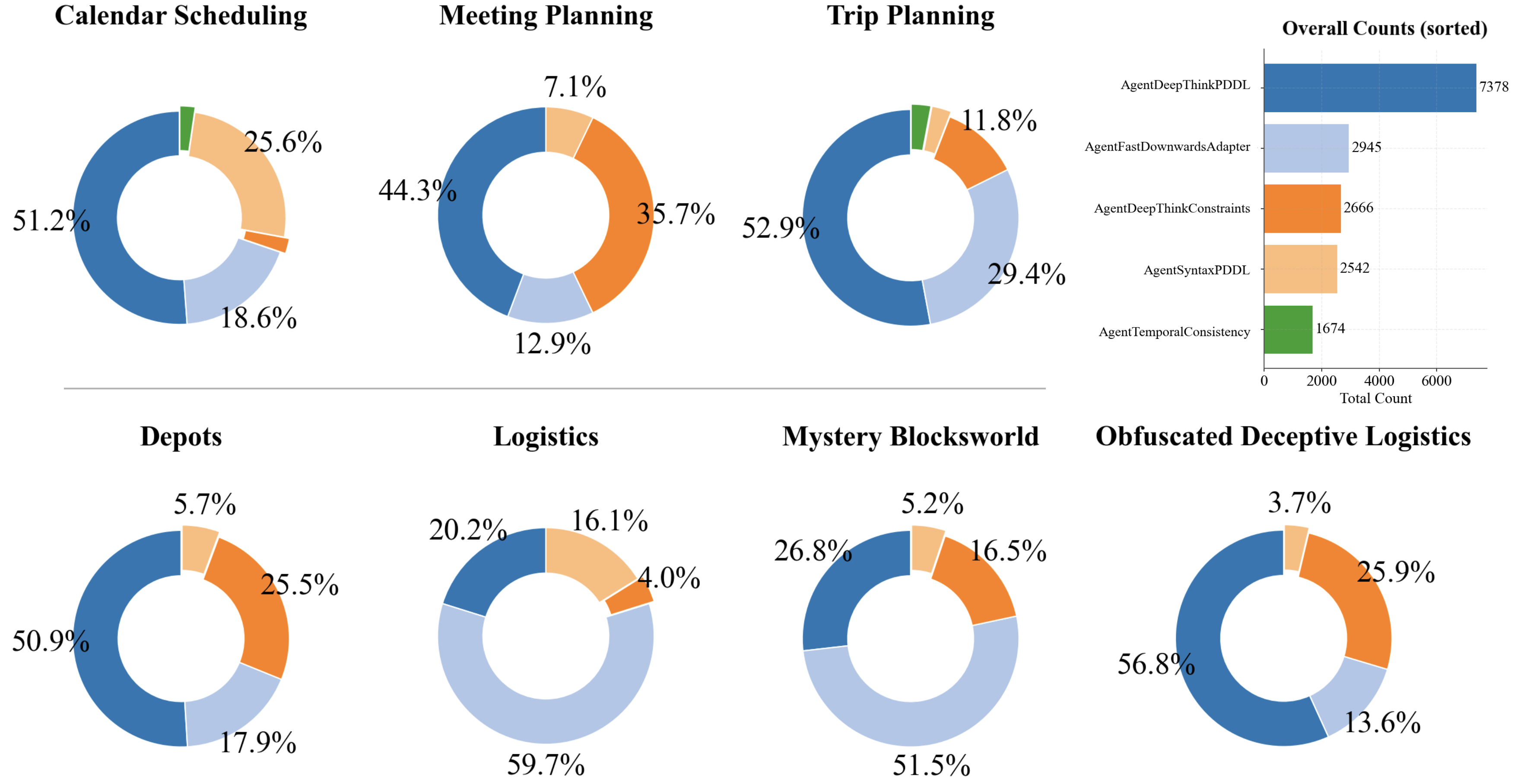} 
\caption{Frequency of hardcoded agents' pickup for GPT-5-mini in the Google Natural Plan Benchmark (top) and Planbench (bottom). On the right, the overall count across the two benchmarks.}\label{fig:frequency-agents}
\end{figure*}

\subsection{On Dynamic and Hardcoded Agents: Efficiency and Agents Choice}

We conduct an ablation study on GPT-5-mini to identify which hard-coded agents are most important for planning. We remove the dynamic agent from the refinement loop and evaluate on the Google Natural Plan benchmark and Planbench. The remaining agents handle PDDL syntax repair, temporal consistency, FastDownward adaptation, and revisiting the original JSON specification to detect missing constraints.

Removing the dynamic agent has little impact on success rates (remaining comparable to Table~\ref{tab:google-planbench}), but increases the number of agent invocations per problem. For instance, with Gemini-2.5-flash on the Google Natural Plan benchmark, the average rises from $3.2$ to $4$, nearly exhausting the budget. A similar trend holds for Planbench, increasing from $2.2$ to $2.8$.

When all agents are available, hard-coded and dynamic agents are invoked with similar frequency (around $50\%$ each). Without dynamic, Figure~\ref{fig:frequency-agents} shows that the orchestrator most often selects the PDDL syntax repair agent. Temporal consistency is frequently used in calendar scheduling, the FastDownward adapter in meeting planning, and the specification-revision agent in trip planning. Planbench follows similar patterns, with syntax repair most common, followed by the FastDownward adapter (depots, obfuscated deceptive logistics) and specification revision (logistics, mystery blocksworld).

\subsection{Discussion: On Benchmarking LLMs for Planning}\label{sec:discussion}
In this section, we highlight key issues in benchmarking LLMs for planning tasks. Beyond informing our results, these observations call for improved methods for reasoning and planning under constraints, and provide empirical support for recent position papers advocating more rigorous evaluation~\cite{katz2025makeplanningresearchrigorous}.

\paragraph{Planning benchmarks are ambiguous, if not wrong.}
In the Google Natural Plan benchmark, we identified several issues with the ground truth labels. Both our method and vanilla models perform well on calendar scheduling but poorly on trip planning, prompting manual inspection. In calendar scheduling, some instances admit multiple valid solutions despite a single labeled answer. In trip planning, the ground truth ambiguously counts travel days as time spent in both locations, which can be unrealistic (e.g., long trips). GPT-5-mini and GPT-5 instead count travel as time spent only in the origin location, leading to lower measured accuracy.
In contrast, Planbench provides well-defined tasks and solutions, though it focuses on classic planning domains (e.g., logistics, depots).

We thus raise the attention of the research community to benchmarks that, if not inspected manually, can mislead our interpretation of LLMs' planning capabilities. We advocate for benchmarks rigorously validated by humans before release. This opens up to the discussion on LLMs as-a-judge to evaluate the correctness of a plan when the ground truth is not available.

\paragraph{The issues of LLMs as-a-judge.}
Last but not least, we want to emphasize how difficult (and potentially, ill-posed) it is to use an LLM-as-a-judge to evaluate the correctness of a plan. As discussed in other works~\cite{gu2025surveyllmasajudge}, unless an LLM has performance that is proven to be superior to that of the model they judge, there is a risk that the judge itself cannot solve the task, and thus understand whether a solution is correct or wrong. Finding better ways to evaluate a plan (e.g., asking a model to output a formatted solution that is tested against the ground truth), is an open problem in planning as well as benchmarking~\cite{lamalfa2023languagemodelsserviceoverview} the research community is still addressing.

To quantify the uncertainty of the LLM as-a-judge, we run three evaluations on the same set of results of Planbench and the Google Natural Plan benchmark, using, as a planner, GPT-5.4 (see Table~\ref{tab:google-planbench}), and, as a judge, GPT-5-mini.
We then measure the variation between evaluations (the difference between max and min success rate). 
As we report in Table~\ref{tab:llm-unc}, the max variation in the success rate for the vanilla is $12\%$ (meeting planning) with an average variation of $6.8\%$ across the seven tasks; for our technique, the max variation is $10\%$ (depots), with an average $3.7\%$ variation across the seven tasks. 

\begin{table}
    \resizebox{\linewidth}{!}{
        \begin{tabular}{lcccc|ccc|c}
        \toprule
        \textbf{} 
        & \multicolumn{4}{c|}{\textbf{Planbench}} 
        & \multicolumn{3}{c|}{\textbf{Google Natural Plan}} 
        & \textbf{Avg. Unc.} \\
        \textbf{} 
        & {\small\textbf{\shortstack{Depots}}} 
        & {\small\textbf{\shortstack{Logistics}}} 
        & {\small\textbf{\shortstack{Mystery \\ Blocksworld}}} 
        & {\small\textbf{\shortstack{Obfuscated Dec. \\ Logistics}}} 
        & {\small\textbf{\shortstack{Calendar \\ Scheduling}}} 
        & {\small\textbf{\shortstack{Meeting \\ Planning}}} 
        & {\small\textbf{\shortstack{Trip \\ Planning}}} 
        & \\
        \midrule
        GPT-5.4 & & & & & & & & \\
        \quad + \textsc{Vanilla} 
        & $6.0$ 
        & $10.0$ 
        & $6.0$ 
        & $7.0$
        & $5.2$ 
        & $12.0$ 
        & $2.0$ 
        & $6.8\pm3.1$ \\
        \quad + \cellcolor{mygrey}\textsc{Agents (Ours)} 
        & \cellcolor{mygrey}$10.0$ 
        & \cellcolor{mygrey}$0.0$ 
        & \cellcolor{mygrey}$0.0$ 
        & \cellcolor{mygrey}$0.0$ 
        & \cellcolor{mygrey}$2.0$ 
        & \cellcolor{mygrey}$8.0$ 
        & \cellcolor{mygrey}$6.0$ 
        & \cellcolor{mygrey}$3.7\pm3.7$ \\
        \bottomrule
        \end{tabular}
    }
    \caption{Uncertainty estimated as the min-max range across same runs with GPT-5-mini as-a-judge.}
    \label{tab:llm-unc}
\end{table}

\section{Conclusion and Future Work}
We introduce a fully automated, end-to-end multi-agent planning framework that combines natural language understanding, dynamic orchestration, symbolic reasoning, and plan interpretation. Our approach enables LLM-based agents to generate workflows, iteratively refine PDDL, and leverage solvers to handle complex planning tasks, with outputs translated back into natural language for interpretability.

Across ten benchmarks (e.g., Google Natural Plan, Planbench, Blocksworld, Hanoi, Sokoban), our method improves success rates on long-horizon planning tasks, while offering smaller gains on CSPs. It also provides a flexible paradigm for handling ambiguity, hallucinations, and coordination via both hard-coded and dynamic agents.

Future work includes scaling orchestration, incorporating multimodal inputs, and applying the framework to real-world and embodied settings, as well as developing theoretical guarantees for dynamically generated workflows.

\section*{Acknowledgments}
SM was supported by the EPSRC Centre for Doctoral Training in Autonomous Intelligent Machines and Systems n. EP/Y035070/1, in addition to Microsoft Ltd. MW was supported by an AI 2050 Senior Fellowship from the Schmidt Sciences Foundation. ELM and SM are affiliated with the Institute for Decentralized AI, which they thank for its support.

\bibliographystyle{abbrv}
\bibliography{bibliography}

\clearpage 
\section{Appendix}

\subsection{The Agents in the Agentic Framework}\label{a:agents}
We implemented several agents, each specialised in a specific task. The core implementation of our framework stands on the following agents.
\begin{itemize}
    \item FastDownwardsAdapter: it adapts the current PDDL domain and problem to be compliant with the syntax of a specific solver (e.g., FastDownwards).
    \item DeepThinkPDDL: it identifies inconsistencies between the constraints, the goal, and the final plan.
    \item DeepThinkConstraints: it focuses on the constraints expressed in the natural language prompt and the JSON representation, to see if they match the PDDL domain and problem.
    \item SyntaxPDDL: it ensures that the PDDL domain and problem are syntactically correct and can be executed by the solver. It specifically focuses on the output of the validator (e.g., uVAL).
    \item TemporalConsistency: it assesses whether the events/actions in the PDDL problem are temporally consistent with the input prompt and the JSON representation.
    \item AgentNaturalLanguage: it turns the final plan into natural actions and the response for the user.
    \item NoOpAgent: it recognises the task has been solved in advance (i.e., without using the whole budget), and terminates the computation.
\end{itemize}

\begin{figure*}[!htbp]
\centering
\includegraphics[width=0.8\textwidth]{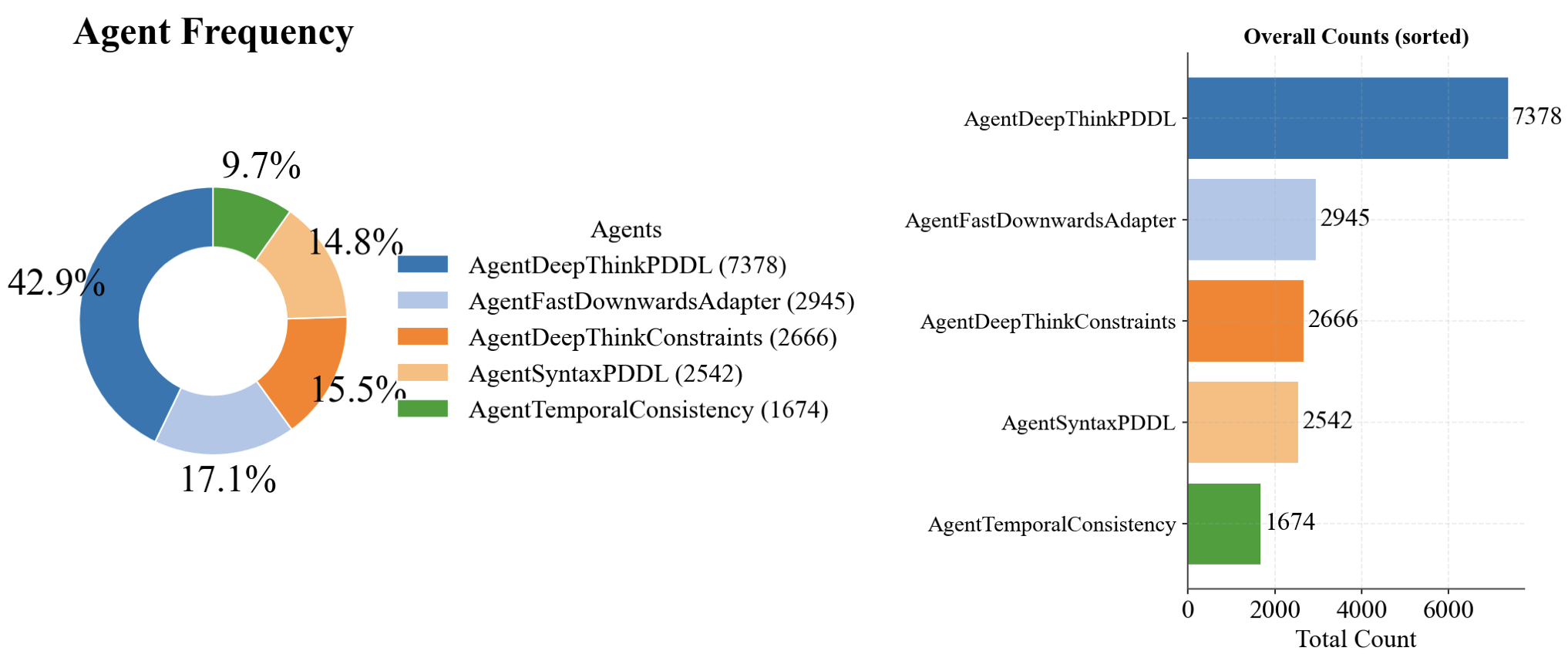} 
\caption{The overall frequency of each agent in the framework, across the Google Natural Language Benchmark and Planbench.}\label{fig:overall-agents-usage}
\end{figure*}

An overall overview of their usage in the Google Natural Plan Benchmark and Planbench is reported in Figure~\ref{fig:overall-agents-usage}. 

Other agents that are effective in boosting the success rate of the model in particular scenarios, but are less employed in our experimental evaluation, are:
\begin{itemize}
    \item AgentAsynchronicity: it optimises the plan so that asynchronous actions are executed at the same time step, if possible. If the solver does not support asynchronous actions, it defines a dummy time variable that allows for async actions (though that is not directly employed by the solver). 
    \item AgentEnforceMultiAgency: it checks whether the PDDL domain and plan correctly implement the specification as a multi-agent system (this is useful when the specification requires multiple actors to perform async actions).
    \item AgentHallucinations: it detects and fixes hallucinations (arising from the LLM, usually) in the authored PDDL domain and problem.
\end{itemize}

\subsection{The Agentic Orchestrator}
The prompt used to call the orchestrator consists of a system prompt that instructs the LLM with the general task to solve, and the prompt, which includes instructions on how to interpret the PDDL logs to call the next agent, alongside a description of each agent's capabilities.
\newline \newline
\textbf{System Prompt:}
\newline
\begin{lstlisting}[language=json]
    You are an orchestrator that coordinates multiple specialised agents. Each agent has a defined role. Read their descriptions, examine the current planning artefacts, and decide which agent is best suited to act next.
    Always discard abstract classes or classes with abstract methods and report the selected class name between <class></class> tags.
    Fast Downward only supports classical STRIPS features; reject or repair PDDL that introduces unsupported requirements such as :fluents, axioms, or conditional effects.
    Pay special attention to artefacts that defer constraints through bookkeeping tokens, post-hoc penalties, or lax goal conditions. Prefer agents that encode the constraints directly.
\end{lstlisting}

\textbf{Prompt:}
\newline
\begin{lstlisting}[language=json,firstnumber=1]
    You are choosing the next agent to improve the current PDDL artefacts.

        Human specification of the task:
        <human_specification>{human_specification}</human_specification>

        JSON plan specification of the task:
        <specification>{specification}</specification>

        Current PDDL domain:
        <domain>{pddl_domain}</domain>

        Current PDDL problem:
        <problem>{pddl_problem}</problem>

        Previously hypothesised solution. Confirm or revise as needed. Think about this carefully!
        <proposed_solution>{proposed_solution}</proposed_solution>

        Plan produced for the {target_solver} solver (may be empty):
        <plan>{pddl_plan}</plan>

        Logs from the solver execution:
        <logs>{pddl_logs}</logs>

        Validator feedback (may be empty):
        <errors>{syntax_errors}</errors>

        Available agents and their capabilities (exclude abstract classes):
        <agents>{agents}</agents>

        Agents previously selected (helps maintain diversity):
        <history>{history}</history>

    Consider selecting the agent for refinement, depending on the issues that you see in the artefacts:
    1. Constraints: ensure the domain and problem satisfy every agent's constraints.
    2. Syntax: enforce compliance with the target solver's PDDL requirements.
    3. Temporal/Causal coherence: disallow token-based bookkeeping, enforce explicit ordering, and prevent oscillating transitions when the specification expects contiguous or monotonic progress.

    Select the class best positioned to address the most pressing issue given the current artefacts.
    Only select the agent <class>NoOpAgent</class> when every objective above is demonstrably satisfied.
\end{lstlisting}

\textbf{dynamic Agent:}
\newline
The content of the tag \{task\_profile\["task\_profile\_prompt"\]\} is generated on the fly by the orchestrator, making this a dynamic agent.
\newline
\begin{lstlisting}[language=json]
You are JackOfAllTrades, the semantic dynamic agent for planning.
Your role is to repair the current PDDL domain and problem by rethinking the planning abstraction for this
specific task instance while staying compatible with {self.prompt_args.get("target_solver", "the target solver")}.

Use the task profile below as weak guidance derived from the problem text:
{task_profile["task_profile_prompt"]}

Behave as the semantic model designer:
- Infer or revise the latent planning abstraction.
- Fix task-specific modeling errors in objects, types, predicates, invariants, action schemas, init, and goal.
- Preserve hard constraints exactly unless the human task explicitly asks for repair.
- If the instance is inconsistent, encode that fact conservatively instead of silently changing the task.
- Use lowercase underscore identifiers consistently.
- Stay within classical solver-compatible PDDL only.
\end{lstlisting}

\subsection{The JSON Representation}
The orchestrator generates, with a specialised module, the JSON representation of the environment, including the agents, their goals, and their constraints.
The JSON file makes explicit the information that is available to each agent, and decides the order of execution to generate the first PDDL domain and problem, which are then fed to the solver and calidator to output the SAS plan.
The JSON representation generated by GPT-5-mini for the Google Natural Plan calendar meeting is the following (all the others are available in the code folder ``environment/static''):
\newline
\begin{lstlisting}[language=json,firstnumber=1]
{
    "name": "CalendarSchedulingExample0",
    "author": "Human",
    "agents": {
        "number": 5,
        "names": [
            "michelle",
            "steven",
            "jerry",
            "auditor",
            "orchestrator"
        ],
        "michelle": {
            "private_information": [
                "Busy on Monday 11:00-12:00"
            ],
            "goal": "Provide Michelle's accurate free time windows within the work hours so a one-hour meeting with Steven and Jerry can be scheduled."
        },
        "steven": {
            "private_information": [
                "Busy on Monday 09:00-09:30",
                "Busy on Monday 11:30-12:00",
                "Busy on Monday 13:30-14:00",
                "Busy on Monday 15:30-16:00"
            ],
            "goal": "Provide Steven's accurate free time windows within the work hours so a one-hour meeting with Michelle and Jerry can be scheduled."
        },
        "jerry": {
            "private_information": [
                "Busy on Monday 09:00-09:30",
                "Busy on Monday 10:00-11:00",
                "Busy on Monday 11:30-12:30",
                "Busy on Monday 13:00-14:30",
                "Busy on Monday 15:30-16:00",
                "Busy on Monday 16:30-17:00"
            ],
            "goal": "Provide Jerry's accurate free time windows within the work hours so a one-hour meeting with Michelle and Steven can be scheduled."
        },
        "auditor": {
            "private_information": [
                "I audit temporal and causal consistency, remove bookkeeping shortcuts such as quota tokens or post-hoc penalties, and ensure all intervals are represented explicitly.",
                "I produce cleaned, normalized availabilities and highlight any implicit assumptions in inputs."
            ],
            "goal": "Audit the provided availabilities for temporal consistency and return a cleaned, explicit availability set for each participant along with notes about corrections or assumptions."
        },
        "orchestrator": {
            "private_information": [],
            "goal": "Consume participants' availabilities and the auditor report and produce a final PDDL domain and PDDL problem targeting the FastDownwards solver that schedules a contiguous one-hour meeting within the work hours satisfying all calendars."
        }
    },
    "environment": {
        "init": {
            "day": "Monday",
            "work_hours": [
                "09:00",
                "17:00"
            ],
            "meeting_duration": "01:00",
            "time_granularity_minutes": 30
        },
        "public_information": [
            "A one-hour meeting must be scheduled on Monday between 09:00 and 17:00.",
            "Meeting participants: Michelle, Steven, Jerry.",
            "Time granularity for proposed slots is 30 minutes and meetings must be contiguous.",
            "There exists at least one feasible one-hour slot that works with all participants' existing schedules."
        ]
    },
    "workflow": {
        "michelle": {
            "provide_availability": {
                "input": [],
                "output": "availability_michelle",
                "system_prompt": "You are Michelle's calendar agent. Using your private busy times and the environment work hours, produce Michelle's free time intervals on Monday within 09:00-17:00. Provide a concise machine-readable list of intervals in 24-hour HH:MM format as an array of pairs, for example: [[09:00,11:00],[12:00,17:00]]. Do not produce PDDL."
            }
        },
        "steven": {
            "provide_availability": {
                "input": [],
                "output": "availability_steven",
                "system_prompt": "You are Steven's calendar agent. Using your private busy times and the environment work hours, produce Steven's free time intervals on Monday within 09:00-17:00. Provide a concise machine-readable list of intervals in 24-hour HH:MM format as an array of pairs. Do not produce PDDL."
            }
        },
        "jerry": {
            "provide_availability": {
                "input": [],
                "output": "availability_jerry",
                "system_prompt": "You are Jerry's calendar agent. Using your private busy times and the environment work hours, produce Jerry's free time intervals on Monday within 09:00-17:00. Provide a concise machine-readable list of intervals in 24-hour HH:MM format as an array of pairs. Do not produce PDDL."
            }
        },
        "auditor": {
            "audit": {
                "input": [
                    "availability_michelle",
                    "availability_steven",
                    "availability_jerry"
                ],
                "output": "audit_report",
                "system_prompt": "You are an expert auditor of temporal constraints. Receive the three participants' availability lists. Normalize times to the environment granularity, remove any implicit bookkeeping shortcuts (for example, blackboxed quota tokens or implicit overlapping allowances), ensure no busy interval overlaps contradict the stated busy times, and produce a cleaned availability structure for each participant. Output a JSON-like object with keys 'cleaned_availabilities' mapping participant name to list of intervals and 'notes' listing any corrections or assumptions made. Do not produce PDDL."
            }
        },
        "orchestrator": {
            "pddl": {
                "input": [
                    "availability_michelle",
                    "availability_steven",
                    "availability_jerry",
                    "audit_report"
                ],
                "output": "pddl_orchestrator",
                "system_prompt": "You are an expert at multi-agent PDDL and the FastDownwards solver. Your task is to produce a PDDL domain and a PDDL problem that, when solved with FastDownwards, finds a contiguous one-hour meeting time on Monday between 09:00 and 17:00 that is within all participants' cleaned availabilities. Use temporal PDDL features or explicit time-slot encoding as appropriate for FastDownwards. Keep participant actions distinct and model the meeting as a single scheduling action with duration 1 hour. Always enclose the PDDL domain between <domain></domain> tags and the PDDL problem between <problem></problem> tags. The domain and problem must be complete and self-contained for FastDownwards. If any input availability is partial or ambiguous, use only the audited cleaned availabilities and do not assume additional free time."
            },
            "prompt": "You are the orchestrator. Use the following public environment information: {environment->public_information} Use participants' availabilities: {availability_michelle}, {availability_steven}, {availability_jerry} and the auditor's cleaned output: {audit_report}. Produce a PDDL domain and a PDDL problem suitable for FastDownwards that schedules the one-hour meeting for Michelle, Steven and Jerry. Enclose the domain between <domain></domain> and the problem between <problem></problem>."
        },
        "constraints": [
            "michelle.provide_availability->auditor.audit",
            "steven.provide_availability->auditor.audit",
            "jerry.provide_availability->auditor.audit",
            "michelle.provide_availability->orchestrator.pddl",
            "steven.provide_availability->orchestrator.pddl",
            "jerry.provide_availability->orchestrator.pddl",
            "auditor.audit->orchestrator.pddl"
        ]
    }
}
\end{lstlisting}

\subsection{Planning Copilot}
\begin{figure*}[!htbp]
\centering
\includegraphics[width=1\textwidth]{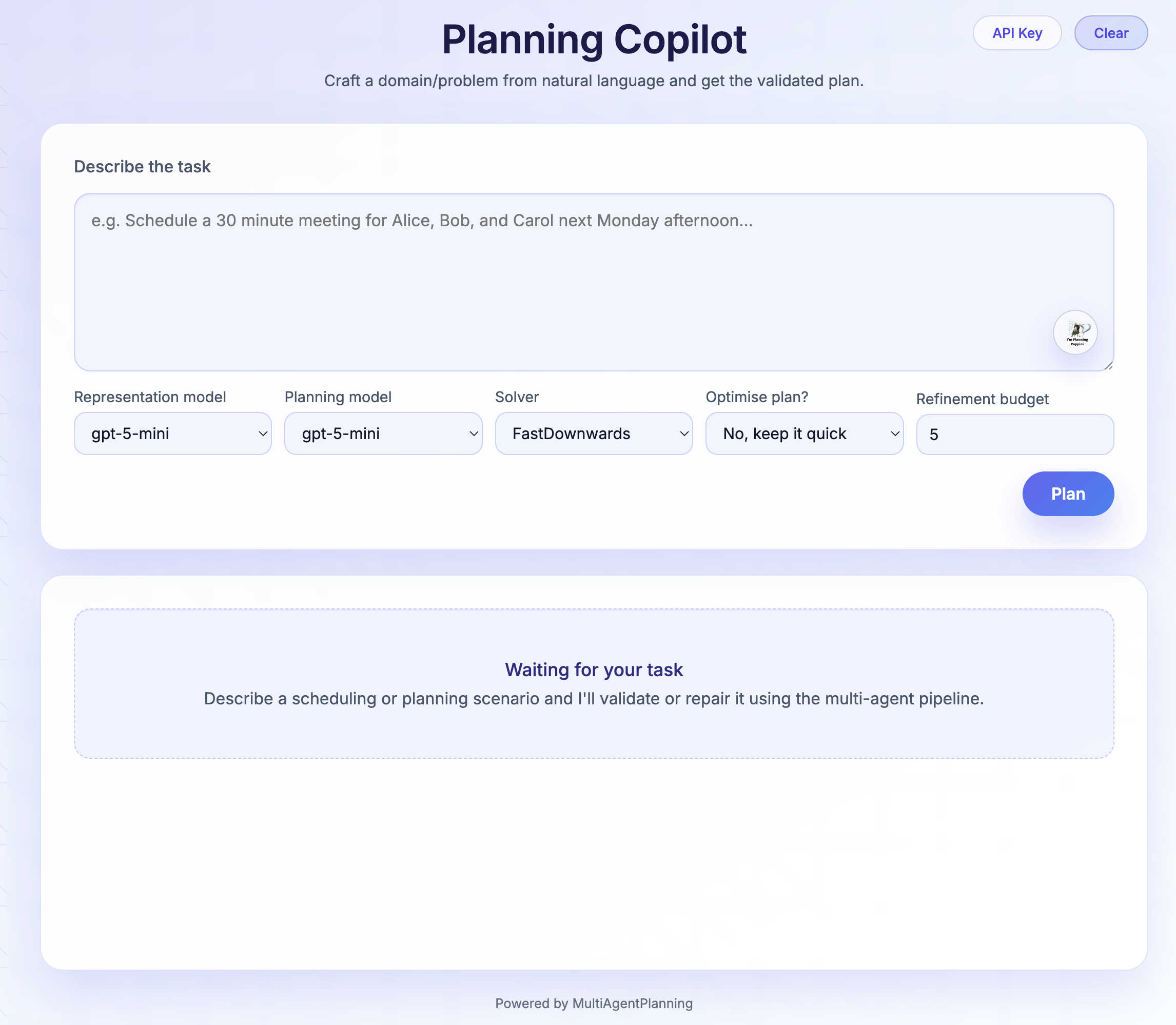} 
\caption{The interface of the Planning Copilot.}\label{diag:webapp-1}
\end{figure*}

\begin{figure*}[!htbp]
\centering
\includegraphics[width=1\textwidth]{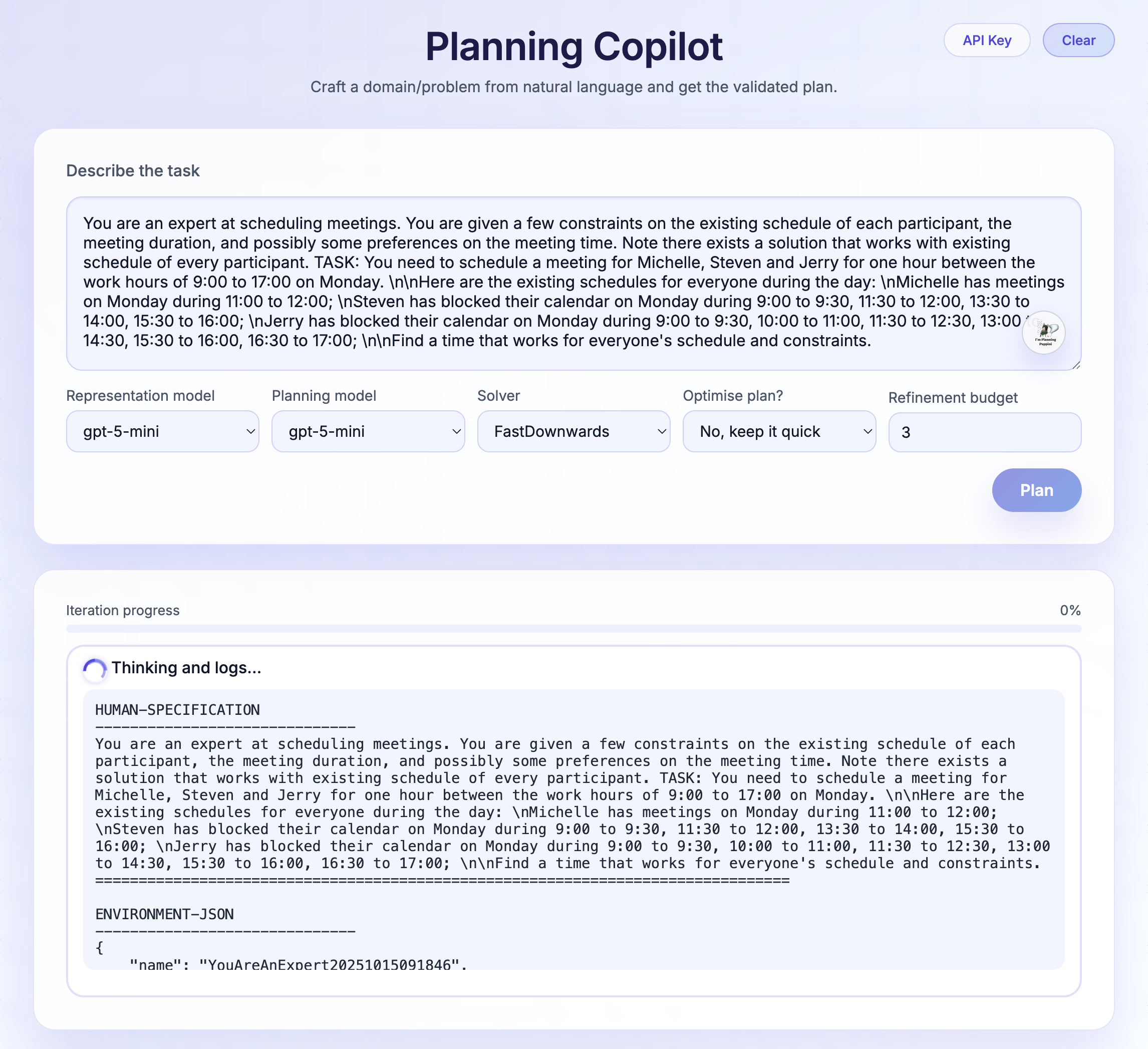} 
\caption{The framework produces the first JSON representation and PDDL domain and problem.}\label{diag:webapp-2}
\end{figure*}

\begin{figure*}[!htbp]
\centering
\includegraphics[width=1\textwidth]{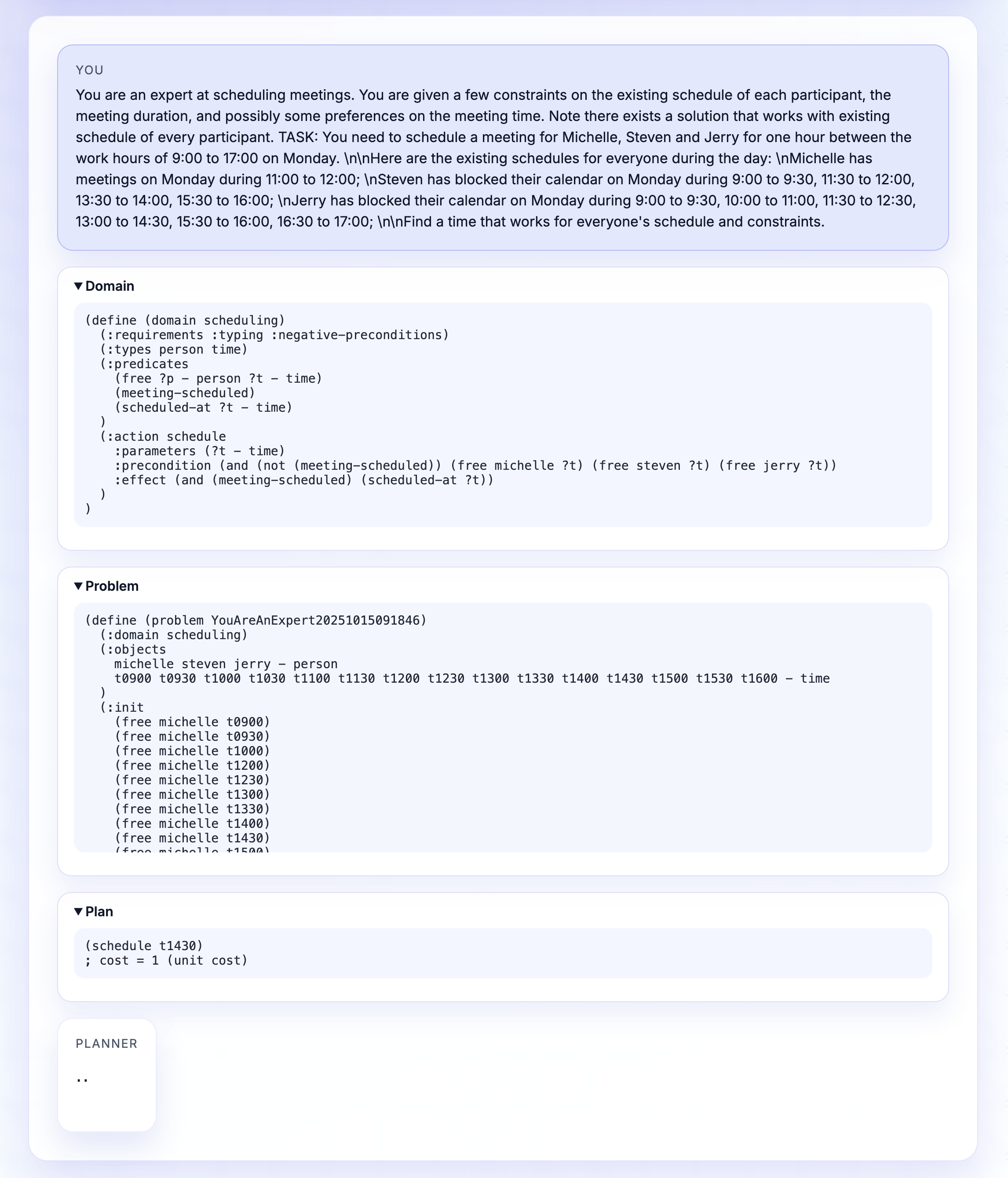} 
\caption{The framework displays the current SAS plan.}\label{diag:webapp-3}
\end{figure*}

\begin{figure*}[!htbp]
\centering
\includegraphics[width=1\textwidth]{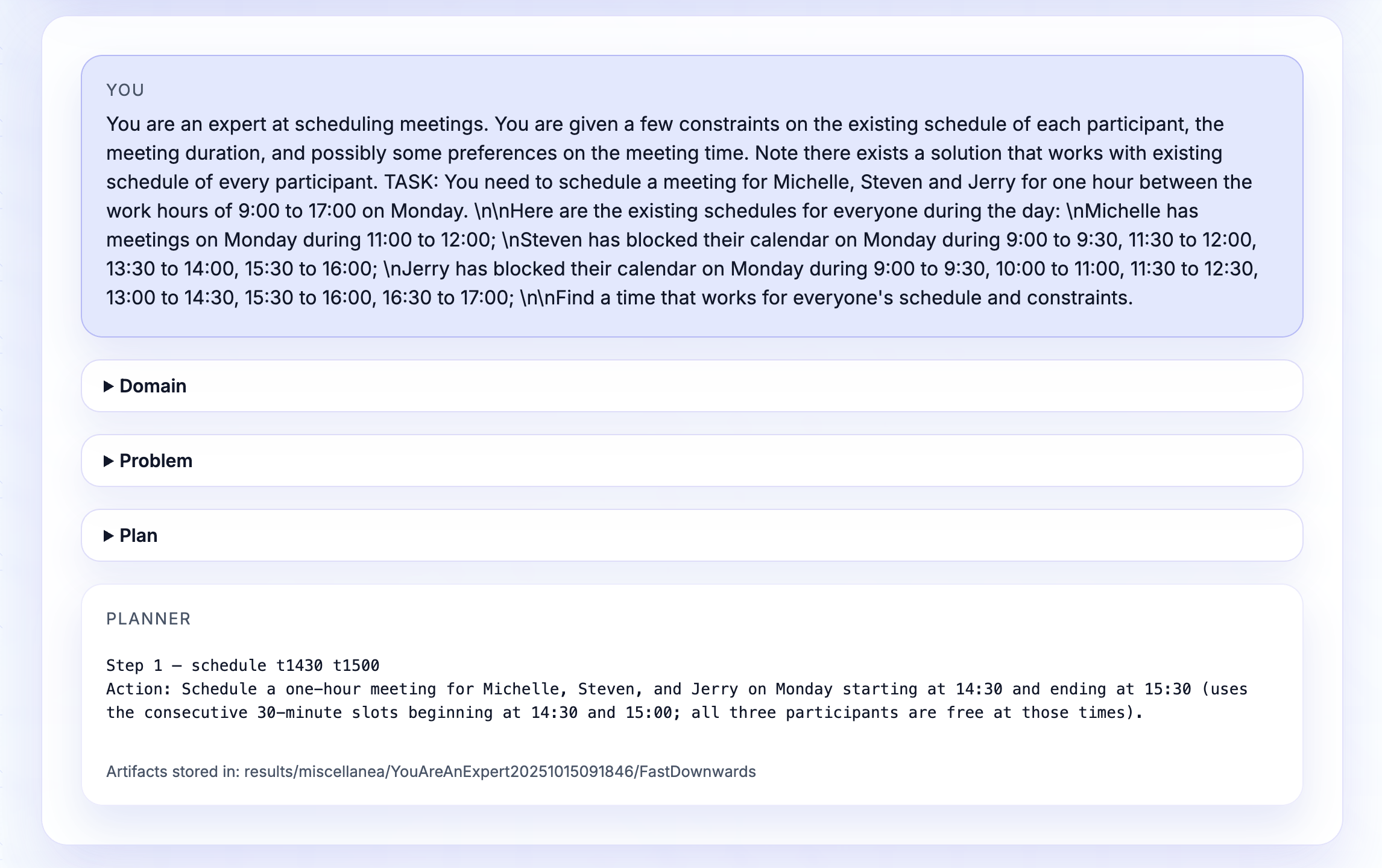} 
\caption{The final SAS plan is back-translated into natural language.}\label{diag:webapp-4}
\end{figure*}

To activate the Planning Copilot, it is sufficient to download the webapp from the code extension and run ``setup.py''. 
Then, by navigating to the local url address ``192.168.0.0:500'', the interface will display (Figure~\ref{diag:webapp-1}).
By clicking on ``API Key'', it is possible to insert a valid ChatGPT key and use the app.
The reasoning process and the produced answer to the first example of the Google Natural Plan benchmark is showed in Figures~\ref{diag:webapp-2},~\ref{diag:webapp-3}, and~\ref{diag:webapp-4}.

\end{document}